\documentclass[runningheads]{llncs}
\usepackage{comment}
\usepackage{graphicx}
\usepackage{booktabs}
\usepackage{amssymb}
\usepackage{subcaption}
\usepackage{marvosym}
\usepackage{cleveref}       
\begin{document}
\title{UI Layers Group Detector: Grouping UI Layers via Text Fusion
and Box Attention}

\titlerunning{UI Layers Group Detector}
%

\author{Shuhong Xiao\inst{1} \and
Tingting Zhou\inst{4} \and
Yunnong Chen\inst{1} \and
Dengming Zhang\inst{2} \and
Liuqing Chen\inst{1,3}\textsuperscript{\Letter}\and
Lingyun Sun\inst{1,3}\and
Shiyu Yue\inst{4}}
\authorrunning{S. Xiao et al.}
%
\institute{
Zhejiang University, Hangzhou 310027, China\and
Chongqing University of Posts and Telecommunications, Chongqing 400065, China\and
Alibaba-Zhejiang University Joint Research Institute of Frontier Technologies, Hangzhou 310027, China\\
\email{ chenlq@zju.edu.cn}
\and
Alibaba Group, Hangzhou 311121, China\\
}

\maketitle              

\begin{abstract}
Graphic User Interface (GUI) is facing great demand with the popularization and prosperity of mobile apps. Automatic UI code generation from UI design draft dramatically simplifies the development process. However, the nesting layer structure in the design draft affects the quality and usability of the generated code. Few existing GUI automated techniques detect and group the nested layers to improve the accessibility of generated code. In this paper, we proposed our UI Layers Group Detector  as a vision-based method that automatically detects images (i.e., basic shapes and visual elements) and text layers that present the same semantic meanings. We propose two plug-in components, text fusion and box attention, that utilize text information from design drafts as a priori information for group localization. We construct a large-scale UI dataset for training and testing, and present a data augmentation approach to boost the detection performance. The experiment shows that the proposed method achieves a decent accuracy regarding layers grouping.

\end{abstract}

\section{Introduction}\label{sec1}
As the central intermediary of human-computer interaction, Graphic User Interface (GUI) is facing great demand with the popularization and prosperity of mobile apps. The traditional process of designing GUI is very long-lasting. It requires investigators to conduct user research, designers to design page materials, and then front-end engineers to code, and it is time-consuming to reach a consensus before multi rounds of back and forth \cite{design-pipeline}.

To achieve faster development and relieve engineers from heavy workloads, some previous researchers applied intelligent methods in automatic GUI generation. Ling et al. \cite{NLP_GUI} considered it a language generation task, and they utilized a language generation method to generate GUI source code from a mixed natural language. REMAUI \cite{reverse} is the first work to introduce GUI screenshots as materials for code generation. They achieved a reverse process from screenshots to GUI code in
 seconds. Pix2code \cite{DBLP:journals/corr/Beltramelli17} then extended the use of images, and they trained an end-to-end deep learning model that generates a source code from a single design input image with 77\% accuracy. More recent commercialization cases like Imgcook \cite{imagecook} construct a platform that allows diversified inputs including screenshots, PSDs, and Figma\footnotemark[1] or Sketch\footnotemark[2] files which contain metadata.

\footnotetext[1]{https://www.figma.com/} 
\footnotetext[2]{https://www.sketch.com/}   


\begin{figure}[htbp]
	\centering
	\begin{minipage}[t]{0.48\textwidth}
	\subcaptionbox{An Example of fragmented layer in UI design.\label{fig1}}{
		\includegraphics[width=160pt,height=60pt]{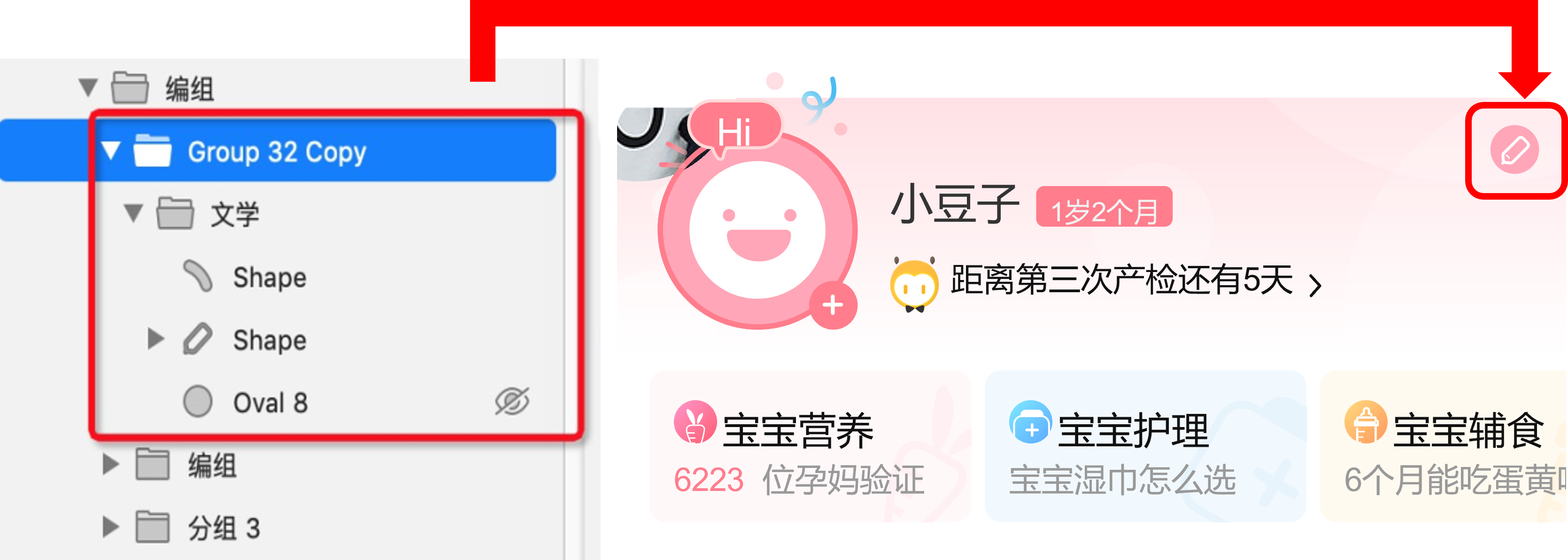}		
	}
	\end{minipage}
    \begin{minipage}[t]{0.48\textwidth}
	\subcaptionbox{Example of grouping layers.\label{fig2}}{
		\includegraphics[width=160pt,height=60pt]{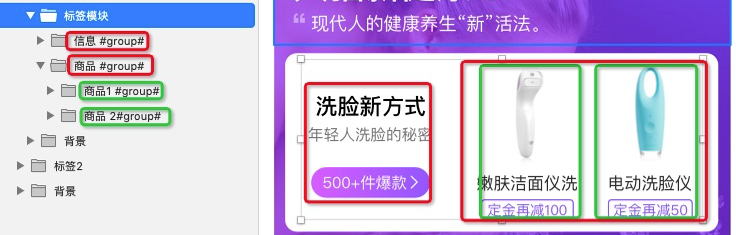}	
	}
	\end{minipage}
	\caption{(a) The icon with the red bounding box is formed with three basic
shape layers. (b) The labeled text and image elements need to be included under the same DOM node using the ``group'' method by adding \#group\# to the target containers.}
	\label{fig(1,2)}
\end{figure}

In the practice of automatic code generation, a massive gap between the design draft (created by digital tools like Sketch) and a quality product (Code and its visual presentation) is that some layers form a whole element in the design draft should be included as a single UI component in the code, while the code generated by automation is hard to reach this. In this case, design drafts should be further constrained by some specific rules to achieve UI code generation with high-quality. For instance, the state-of-art solution, Imgcook, highlights ``merge'' and ``group'' as two of the most important rules that reorganize the structure of design drafts to bridge the gap we described. The ``merge'' method integrates multiple fragmented layers representing basic shapes (e.g., rectangle, oval, path) and visual elements(e.g., text and image) into a single image. As illustrated in Fig. \ref{fig1}, the icon with the red bounding box is formed with three basic shape layers. Without structured merging, these fragmented layers confuse the AI in understanding the semantic meaning of UI components and affect the readability and reusability of generated code. In the other case, the ``group'' method deals with malposed structures in the design draft. As illustrated in Fig. \ref{fig2}, to ensure that generated code does not cause element loss or structural redundancy, the labeled text and image elements need to be included under the same DOM node using the ``group'' method by adding ``\#group\#'' to the target containers. However, Imgcook requires front-end engineers to manually locate the layers to proceed operate the ``merge'' or ``group'' method. 
Manual identification is time-consuming and prone to omissions because of the numerous layers and diverse nested structures. In this paper, we focus on automatically recognizing ``group'' problems. More specifically, we try to locate and group images (we define images as all basic shapes and visual elements) and text layers that have the same semantic meaning in the design draft. Therefore, We can optimize the design layout and obtain high-quality UI code.  




\begin{figure}[htbp]
	\centering
	\begin{minipage}[t]{0.48\textwidth}
	\subcaptionbox{An Example of semantic consistency.\label{fig3}}{
		\includegraphics[width=160pt]{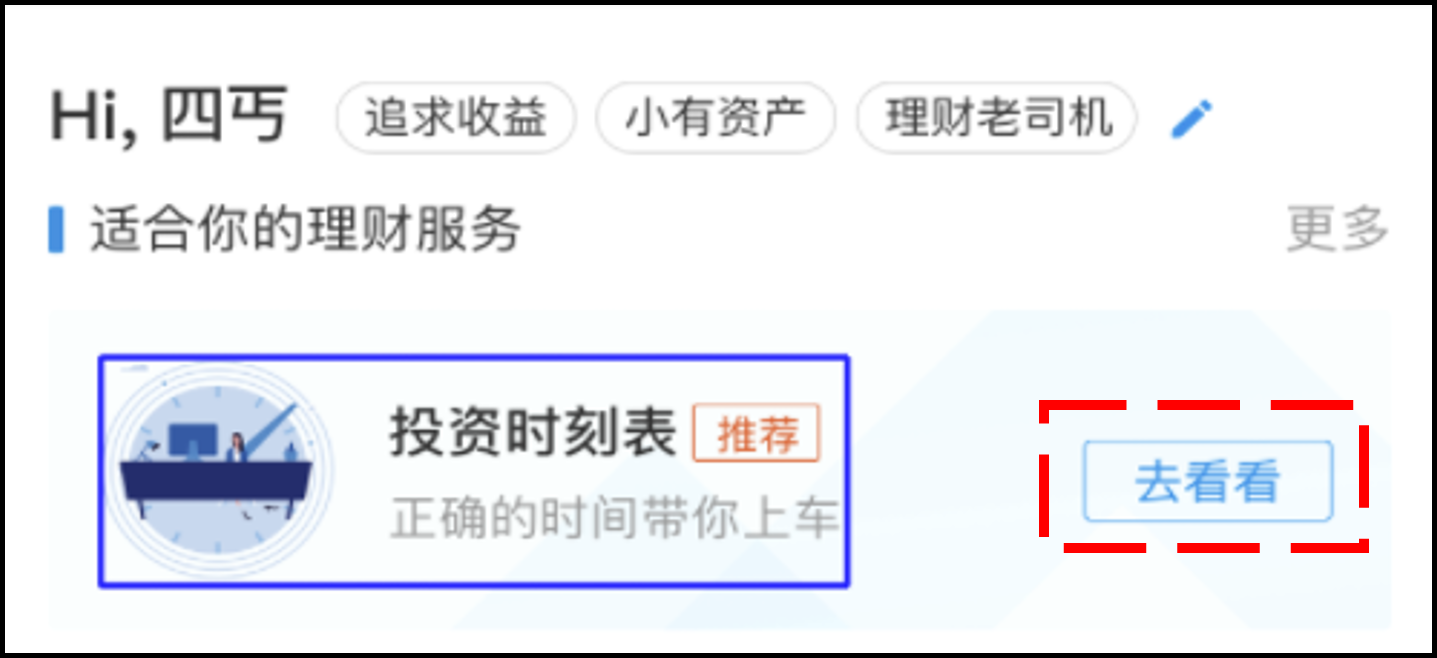}		
	}
	\end{minipage}
    \begin{minipage}[t]{0.48\textwidth}
	\subcaptionbox{An Example of various image-text group patterns\label{fig5}}{
		\includegraphics[width=160pt]{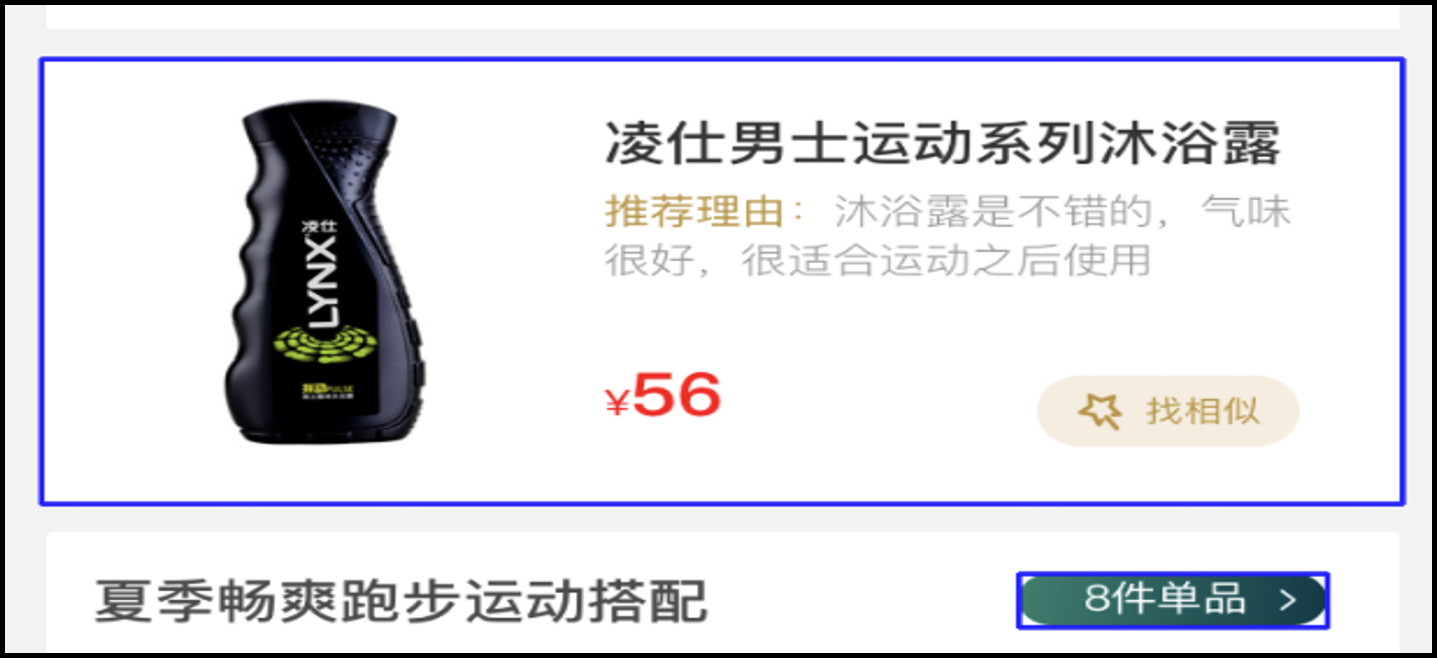}	
	}
	\end{minipage}
	\caption{(a) The clock image in blue box shares a close semantic meaning with the text ``investment schedul'', while the text ``view it'' in the red box represents another meaning. (b) The group strategy is different for the banner and small icon. We keep some background for the banner in order to avoid missing some elements (that maybe invisible) while bound the small icon group as tight as possible.}
	\label{fig(3,4)}
\end{figure}

To address this issue, We proposed our UI Layers Group Detector that utilizes object detection techniques to detect the area to be grouped on real screenshot images. As multi-modal approaches have shown its great power in general UI component detection task\cite{muti}, we introduce text embedding and box attention mechanism that use text-related information as extra modalities to help improve our Detector. The text layers are used as a local semantic focal, together with the global image feature to benefit generated proposals. Given an arbitrary design draft, we follow semantic consistency as the grouping criterion, i.e., not all adjacent visual elements will be integrated into a single group. As illustrated in Fig. \ref{fig3}, the clock image in blue box shares a close semantic meaning with the text ``investment schedule'', while the text ``view it'' in the red box represents another meaning. Under this condition, precisely predicting the group range is challenging because of the background layer and elements around. Another challenge is that the diversified UI application scenarios result in various group patterns with different sizes and element numbers. For example, as illustrated in Fig. \ref{fig5}, for the group of banner contains a thumbnail with text information like product name, price, and description, we have the empty background layer inside the bounding box to better retrieve all the banner elements in the design draft. This is because for group like banner which contains abundant elements, elements sometimes partially invisible so that they will be dropped if we apply a tight bound. While for small icon like the group only contains ``\textgreater'' and text ``8 items'', we make the bounding box as tight as possible so that no irrelevant element will be covered.

In particular, our method solves the two challenges by following approaches.We adopt a state-of-art object detection model to achieve precise localizing of group objectives.To deploy our Group Detector as a plug-in for UI code generator like Imgcook, we choose to adopt from the Faster-RCNN because it is lightweight, delicate, and easy to modify. We collect our dataset from the Sketch files of the most frequently used mobile apps. The target groups on each image are labeled carefully by professional labelers guided by the semantic consistency we described before. The high-quality dataset allows our data-driven model to make more accurate predictions. We summarize our contributions as follows:
\begin{enumerate}
\item We construct a high-quality dataset containing UI screen images from widely used mobile apps. An image augmentation algorithm is introduced to boost the performance of our method.

\item We proposed our UI Layers Group Detector which takes UI screenshots and metadata from design drafts and solve the image-text grouping task. This work is designed to fill in the gap in the automatic UI code generation works.

\item We proposed the text fusion to incorporate features from text layers to the related image region. And the box attention mechanism is introduced as another plug-in component with creating an extra spatial binary image which encodes the position of each text layer inside the UI image. 

\item We carry out experiments on the constructed dataset to verify that the UI Layers Group Detector achieves a decent accuracy regarding UI layers grouping.

\end{enumerate}

\section{Related Works}\label{sec2}
\subsection{Intelligent UI Code Generation}\label{subsec2.1} 
Automation in UI code generation has become an attractive topic after the machine learning and artificial intelligence boom. Early intelligent automation research was mainly used to replace template-based UI design, where users spend time searching for suitable materials to compose their design. We can further classify these works by the level of the fidelity of the design prototype they use as input \cite{auto-survey}. Batuhan et al. \cite{hand-draft} use hand-drawn images to identify and generate basic buttons, text, images, and other components. Works as sketch2code \cite{Sketch2code} utilizes design drafts with more details and achieves the automatic generation of UI structure. Pix2code \cite{DBLP:journals/corr/Beltramelli17} follows a similar approach to generating textual descriptions from photographs. It takes actual UI screenshots as input and achieves a high accuracy generation. More recently, commercial platforms like Imgcook \cite{imagecook} takes advantage of metadata in professionally designed software and achieve a generation with high quality and reusability.

\subsection{UI Design Check}\label{subsec2.2} 
With the popularization of automated UI generation technology, in practical applications, it is necessary to evaluate the generation quality and solve the problems like missing elements or components overlapping caused by hardware or software compatibility. OwlEye \cite{owleyes} detects GUIs with display issues and locates the detailed region of the issue based on the deep learning method. LabelDroid \cite{Unblind} focuses on image-based buttons and achieves a highly accurate prediction of labels by learning from large-scale commercial apps in Google Play. FSMdroid \cite{fsm} uses the MCMC sampling method to analyze GUI apps dynamically and detect defects that reside on unfrequented trails.  

Considering our work as an object detection task, we briefly review the recent advances in this field. Anchor-based approaches like Faster-RCNN \cite{7485869} define a set of anchor boxes of different sizes and use feature maps from different convolutional layers to classify and regress anchor boxes. The following SSD \cite{SSD}, Yolov2 \cite{yolo}, and RetinaNet \cite{retina} continue this idea and achieve a state-of-art performance in natural object detection. Some achievement has also been made in using Object Detection in UI-related areas. Li et al. \cite{data-clean} introduce the CLAY pipeline for UI screen dataset cleaning. They address the mismatches in visual elements and metadata. Zang et al.\cite{muti} leverage object detection on recognizing UI icons and achieve good performance. 


\begin{figure}[htbp]
\centering
\begin{minipage}[t]{0.9\textwidth}
\centering
\includegraphics[width=0.9\textwidth]{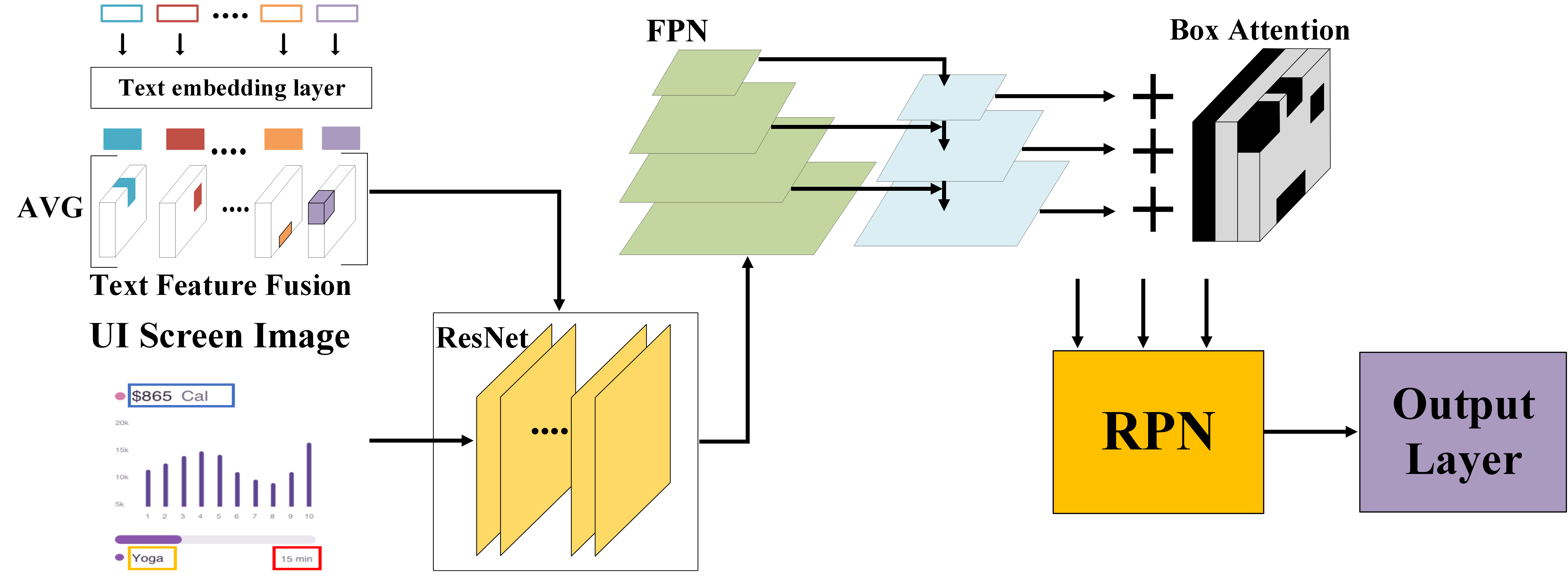}
    \caption{ The overview of proposed method. }
    \label{fig4}
\end{minipage}
\end{figure}

\section{Methods}\label{sec3}
In this section, we introduce the proposed approach to our UI Layers Group Detector described in Section \ref{sec1} as shown in Fig. \ref{fig4}. We first introduce our work on collecting UI screen dataset based on Sketch file, which is widely used in software UI design. Image segmentation is applied considering the nature of small and regional target detection (Section \ref{subsec3.1}). We then introduce our Group Detector via the order of basic setting, text fusion, and box attention. We introduce the strategy of text content and bounding boxes as extra features to help with target localization. The embedding text features are fused into the early convolution layer of our backbone(Section \ref{subsec3.2.1}). We then introduce box attention as another way of utilizing text information. A spatial binary image is created based on text bounding boxes to guide the target detection process by revealing potential image-text group areas. We fuse the box attention with the image feature maps on all FPN output layers and feed them into the RPN for better proposals(Section \ref{subsec3.2.2}). 

\subsection{Dataset}\label{subsec3.1}  

We create a high-quality UI dataset based on UI layouts developed by professional designers using Sketch. Each artboard in a Sketch file represents a UI design for Android or iOS mobile Apps. We eliminate unusual designs that placed the components without artboards as containers and convert the rest as UI screen images. A group of professional labelers was recruited to generate our target group annotation. We adapt our dataset based on COCO-style \cite{https://doi.org/10.48550/arxiv.1504.00325}, including images, annotations, categories, and supplementary text consisting of text semantics and position in the images.  
Considering the nature of the UI screen with a large aspect ratio, while our targets are usually small and regional, image segmentation is applied to all collected UI screens. Given an arbitrary screenshot image, we split it along the long side. To avoid distortion, we keep each piece a square shape, and only bounding boxes that are entirely inside should be recorded. To avoid some bounding boxes being separated into two samples, we leverage a sliding stride to make sure every bounding box will at least appear in one image.

We then split our dataset into training, validation, and test sets for experiments. The split is performed package-wise and image-wise, i.e., artboards in the same Sketch design file and segmented images from the same UI screen image are not shared across different splits. This approach avoids information leakage because UI screen images from the same design might have similar layouts. As a summary, the number of screen images in our dataset increases by a factor of 3 after applying the segmentation. We have 4946 screen images in the training set with 16533 labeled groups and 11617 texts. For the validation set, we have 601 screen images, 2093 labeled groups and 2093 labeled groups. And for the test set, we have 681 screen images, 2410 labeled groups and 1442 labeled groups.

\subsection{UI Layers Group Detector}\label{subsec3.2}   

Usually, the placement and size of our target groups vary widely across the different UI designs. Therefore, the most crucial step in our approach is to infer the bounding boxes accurately. To achieve this goal, we utilize ResNet-50 \cite{resnet} and FPN (Feature Pyramid Networks) \cite{fpn} as our backbone for extracting feature maps. At the RPN (region proposal network), we use a softmax to determine positive or negative anchors and a bounding box regression to modify anchors to obtain precise proposals. Then we borrow the RoI Align \cite{maskrcnn} to replace the RoI Pooling for proposal maps extraction based on input feature maps and proposals. Instead of quantized operations, the RoI Align utilizes linear interpolation to reduce the precision loss. At the classification stage, bounding box regression is introduced again to obtain each proposal's position offset for regressing more accurate target bounding boxes.
\subsubsection{Text Fusion}\label{subsec3.2.1}  

This section introduces the text embedding features as a plug-in component to the Group Detector. As we discussed in Section \ref{sec1}, the grouping patterns varies in different application scenarios. While we also find that components with the same functionality are similar in appearance and should further apply the same grouping strategy. In this paper, together with the pixel-based UI screen image data, we further utilize the text layer information in each Sketch design, which contains the text layer position as bounding boxes and text contents. The supplementary text-level knowledge enables us to identify potential pattern-text groups and reveals the function of target groups in the UI screen.  

Given a UI screen image in our dataset, let $\{bb_i\}_{i=1} ^{N}$ be all text layer bounding boxes inside the image and $\{tt_i\}_{i=1} ^{N}$ be all text contents. Let $I \in \mathbb{R}^{3\times H \times W}$ as input image and $C \in \mathbb{R}^{D\times H' \times W'}$ be an intermediate feature map extracted from the first convolution layer $Conv1$ of ResNet-50. Our first step is to construct a feature map $T_i \in \mathbb{R}^{K\times H \times W}$ for every text information with the same height $H$ and width $W$ using the bound box $bb_i=\{x_{min},y_{min},x_{max},y_{max}\}$. To achieve this, we use a text encoder to transform text content $tt_i$ into corresponding text embedding $e_{i}\in \mathbb{R}^{K\times 1 \times 1}$. We let $T_i$ fill with all 0 initially and update the text embedding $e_i$ inside the bounding box range. Specifically, we set $T_i[:,p,q] = e_i$ where $p\in[y_{min} \times H,y_{max} \times H]$ and $q\in[x_{min} \times W,x_{max} \times W]$. In the next step, we calculate $T = avg(T_i)_{i=1}^N \in \mathbb{R}^{K\times H \times W}$ and apply the same $Conv1$ and an extra $1 \times 1$ convolution layer to get the text feature map $T'$ with the same size of pixel feature map $C$. In the final step, we calculate a new feature map $F = T'\circ C$, where $\circ$ denotes the element-wise addition, and this new fused feature map is fed into the later convolution layer of Resnet-50. 

\subsubsection{Box Attention}\label{subsec3.2.2}  
In this section, we introduce box attention as another way of utilizing text information to improve our Group Detector. We borrow the idea of box attention from Bunian et al \cite{vins}. This mechanism was first introduced by Kolesnikov et al \cite{boxattention}. to model object interactions in a visual relationship detection task. Bunian et al. adopted it in their object detection pipeline. The idea of box attention is to create an extra spatial binary image encoding the position of each text layer inside the UI screen. The binary image here, is considered as a position-focal that together with the global image information to understand compositional relationship among elemets.

Specifically, given an image feature map $F_i \in \mathbb{R}^{D\times H_i \times W_i}$ and text bounding box $\{bb_j\}_{j=1} ^{N}$ where $bb_j = \{x_{min},y_{min},x_{max},y_{max}\}$ corresponding to input UI screen image height $H$ and width $W$, we first transform the bounding box based on $H_i$ and $W_i$ and create the bounding box map $B_{i,j}\in \mathbb{R}^{3\times H_i \times W_i}$. We set $B_{i,j}[0,p,q] = 1$ where $p\in[y_{min} \times H_i,y_{max} \times H_i]$ and $q\in[x_{min} \times W_i,x_{max} \times W_i]$, $B_{i,j}[1,:,:] = 0$ and $B_{i,j}[2,:,:] = 1$. In the next step, we calculate the overall box attention map $B_i$ for the $i^{th}$ image feature map as $B_i = avg(B_{i,j})_{j=1}^N$ and an extra $1 \times 1$ convolution layer is applied on it to match the image feature map dimension $D$. In the final step, we add the box attention map with each FPN output image feature map $F_i$ to get the fused feature map $M_i = F_i\circ B_i$ for further proposal generating.

\section{Experiments}\label{sec4}

\subsection{Implementation Details}\label{subsec4.1}  
We implemente our model using MMDetection \cite{mmdetection} codebase. We use ResNet-50 pre-trained on ImageNet together with the FPN as our backbone network. We set the anchor size [32,64,128,256,512] with anchor ratio [0.5,1.0,2.0,4.0,8.0] for the potential tiny and nonsquare target. The input UI screen images are resized into size in [800,1300] before being fed into Resnet-50. The output features of stage1-4 of ResNet-50 are fed into FPN, together with a further MaxPooling layer, giving us the five feature maps with different scales. For FPN settings, we follow the standard settings as in \cite{fpn}. Proposals are computed from all five pyramid feature maps, and RoI Align is performed with bilinear interpolation.  

For text embedding fusion,  we first encode the text contents into vectors with length $K$ using pre-trained transformers \cite{wolf-etal-2020-transformers}. We set the parameter $K$ to 16, considering the average text length of 11.4. After generating the text feature map $T \in \mathbb{R}^{K\times H' \times W'}$, we increase the channel size from $K$ to $D$ which is the channel size of feature map at Conv1 of stage 0 in ResNet-50. For box attention, we adopt five spatial binary images with the same size as the pyramid feature maps.  

For the training details, we train our model with a mini-batch of 2 for 72 epochs using SGD optimizer with a momentum update of 0.9 and a weight decay of 0.0005; and set the initial learning rate 0.01 with a decay of factor 0.1 every 10 epochs during training. We train our model on an NVIDIA GeForce RTX 3080Ti GPU and it takes about 9 hours for the model to converge.  

\subsection{Evaluation Metrics}\label{subsec4.2} 
In this paper, we report performance metrics used in the COCO detection evaluation criterion \cite{cocometric} and provide mean Average Precision (AP) across various IoU thresholds i.e. IoU={0.50:0.95,0.50,0.75} and various scales: {small, medium and large}. Without further specified, we refer mAP[0.50:0.95]to as AP.  


 \subsection{Results}\label{subsec4.3}
 \textbf{Detection performance after image segmentation.} We first test the benefits of our image segmentation algorithm. As shown in Table \ref{table3}, the image segmentation contributes to a performance boost of AP by about 20\%. In the UI design draft, a potential group like a banner or an icon only takes up a small area of the overall screen. Even if we have modified the anchor's size and aspect ratio accordingly, finding all targets is still challenging. With segmentation, the target becomes more prominent in the background. Furthermore, we also get an augmentation effect with an overlapping slide window that applies multiple locations for each target in image slices.

\begin{table}[htbp]
\centering
\caption{Detection performance after image segmentation}\label{table3}%
\begin{tabular}{@{}llllllll@{}}
\toprule
Method  &$AP$  & $AP_{50}$ & $AP_{75}$&$AP_{S}$  & $AP_{M}$ & $AP_{L}$ \\
\midrule
No segmentation & 0.428   & 0.585  & 0.474 & 0.367&0.408&0.446\\
apply segmentation &  \textbf{0.625}   &  \textbf{0.799}  &  \textbf{0.699} &  \textbf{0.550}& \textbf{0.593}& \textbf{0.648}\\
\bottomrule
\end{tabular}
\footnotetext[1]{All the results are tested using MMDetection}
\end{table}

\textbf{Detection Models comparison.} We then compare the detection performance using recent state-of-art object detection models as shown in Table \ref{table2}. All the results are tested using the segmented dataset. To deploy our Group Detector as a plug-in for UI code generator like Imgcook, we expect it to be lightweight while achieving high accuracy. "UIGD-TF" denotes our UI Layers Group Detector with the text fusion and "UIGD-BA" denotes our UI Layers Group Detector with the box attention. Here we could see that "UIGD-TF" achieves the highest AP performance of 0.658 among all the model we experimented. And "UIGD-BA" takes the second position with AP performance of 0.650. They are also lighter than other models. For example, "UIGD-TF" is 66.9\% lighter on model parameters than the Sparse-RCNN while 3.7\% higher on accuracy.

\begin{table}[htbp]
\centering
\caption{Detection performance comparison}\label{table2}%
\begin{tabular}{@{}lllllllll@{}}
\toprule
Method   & Parameters $\downarrow$ &$AP$  & $AP_{50}$ & $AP_{75}$&$AP_{S}$  & $AP_{M}$ & $AP_{L}$ \\
\midrule
Deformable DETR& 49.76M& 0.603  &0.751    & 0.670  & 0.439 &0.558&0.635\\
YoloX  & 54.15M  & 0.592   &0.788  & 0.654 &0.448&0.517 &0.634  \\
Sparse-RCNN  & 125.21M & 0.621  & 0.759  & 0.683 &0.500 & 0.566 &  0.657 \\
Faster-RCNN & 41.35M & 0.625  & 0.799  & 0.699 &0.550 & 0.593 & 0.648 \\
UIGD-TF & 41.30M&  \textbf{0.658}   &  \textbf{0.853}  &  \textbf{0.729} & \textbf{0.572}&0.643& \textbf{0.708}\\
UIGD-BA &41.29M& 0.650   & 0.825 & 0.715 &  0.568& \textbf{0.651}&0.678\\
\bottomrule
\end{tabular}

\end{table}

\textbf{Performance with text fusion and box attention.} We then investigate the contribution of our proposed text fusion component. As shown in Table \ref{table2}, compare with the Fatser-RCNN, which got the highest accuracy in the rest of the model, we get an 3.3\% AP improvement brought by the text fusion strategy suggests the necessity of the text sematic and location prior knowledge, which enriches the feature representation. The box attention also gives a 2.5\% AP improvement, indicating that concentrating on text location help captures potential group targets. Comparing the model parameters, it can be ssen that apply the text fusion and the box attention only increase the model parameters  by 0.48\% and 0.24\% while the accuracy is significantly increased by 5.28\% and 4.00\%. All the results show that our proposed text fusion and box attention work effectively, and achieve our requirement of lightweight.



\begin{figure}[htbp]
	\centering
	\begin{minipage}[t]{0.48\textwidth}
	\subcaptionbox{Example of good cases.\label{fig7}}{
		\includegraphics[width=160pt,height=110pt]{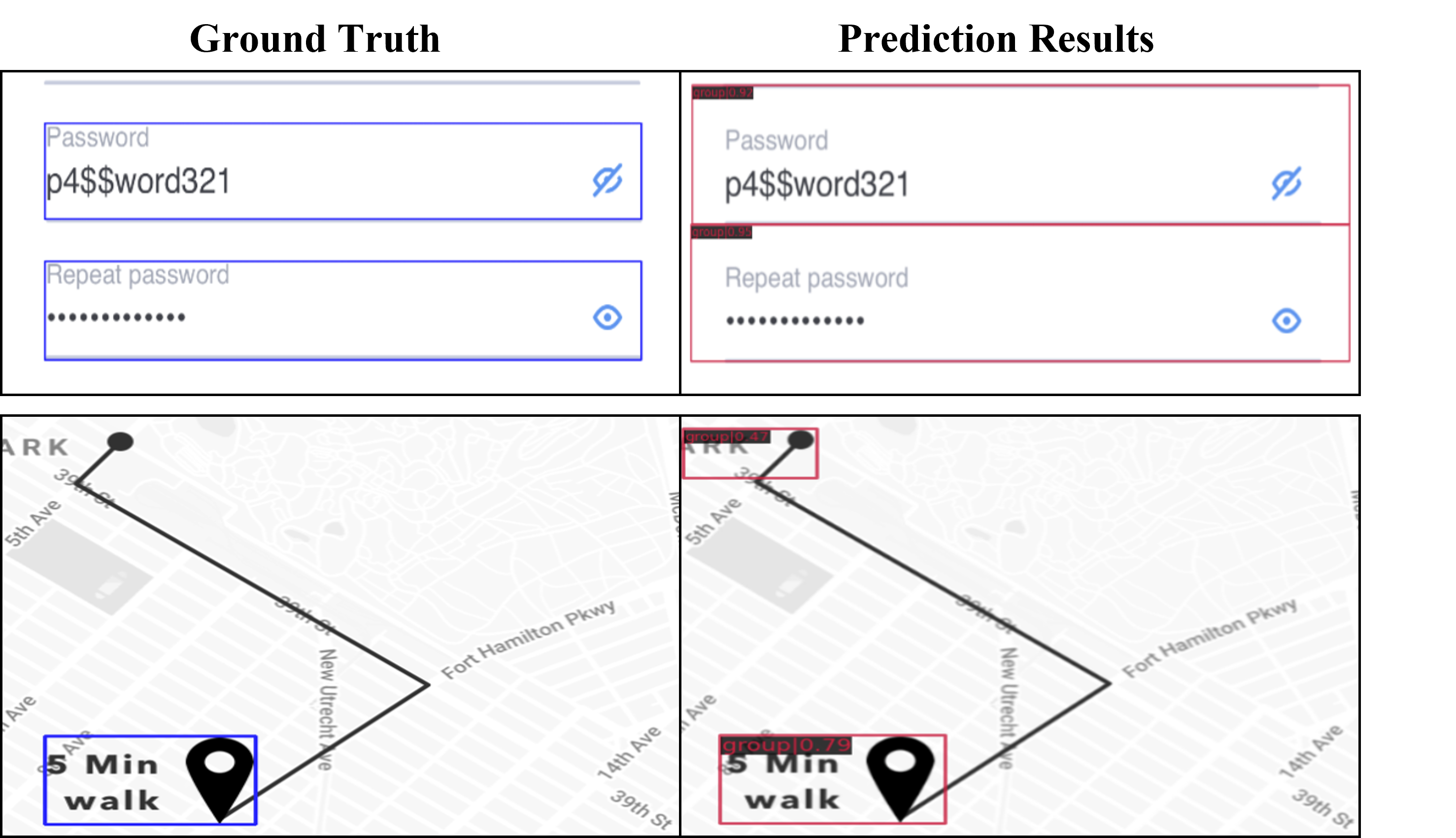}		
	}
	\end{minipage}
    \begin{minipage}[t]{0.48\textwidth}
	\subcaptionbox{Example of failure cases.\label{fig6}}{
		\includegraphics[width=160pt,height=110pt]{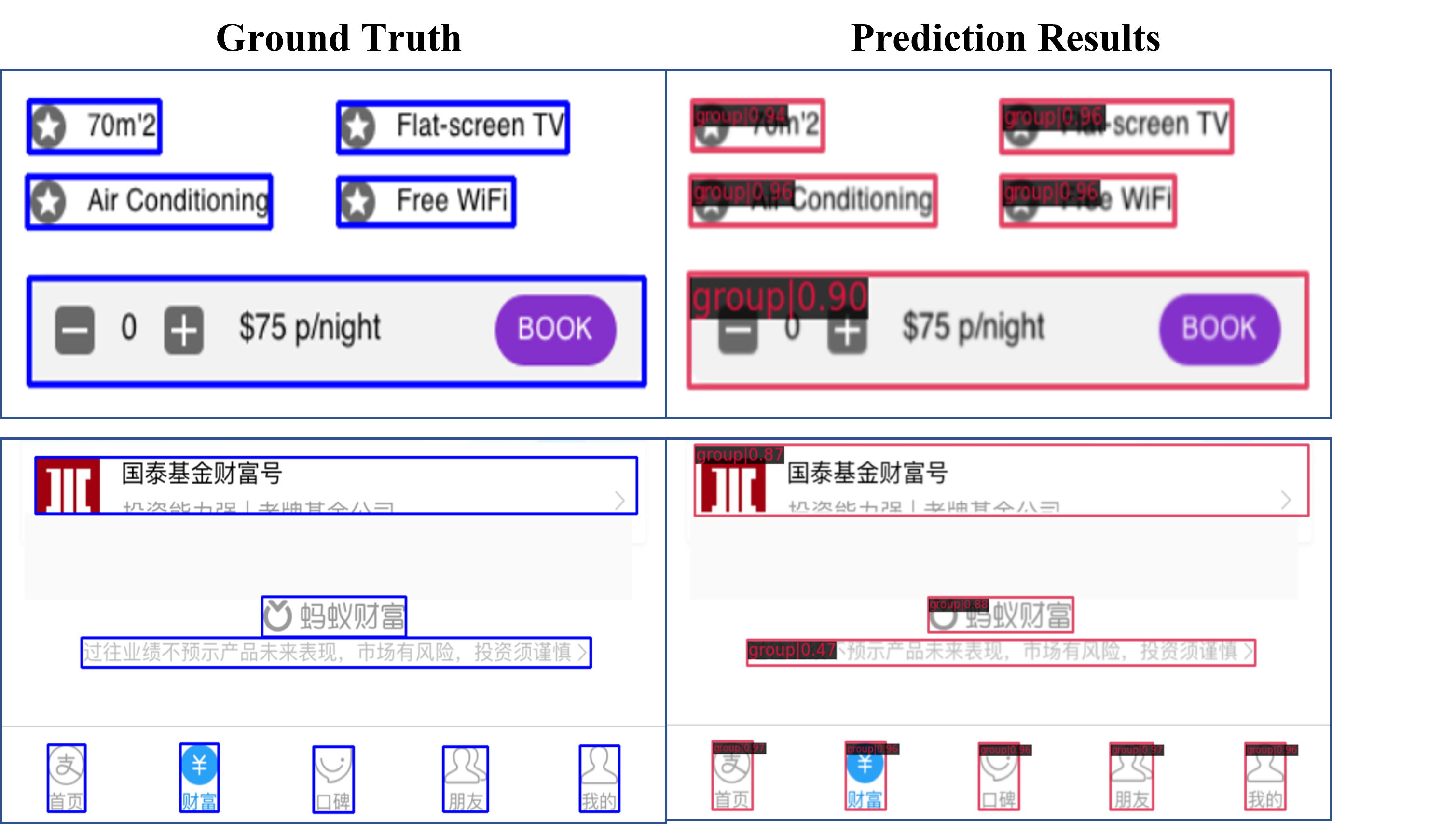}	
	}
	\end{minipage}
	\caption{(a) Our Group Detector successfully localize all the target groups with highly matched bounding boxes. (b) Our Group Detector show weakness on perfectly fit the ground truth bounding boxes on empty background around, and produces the wrong groups when faced with complex multi-level structures. }
	\label{fig(6,7)}
\end{figure}

\textbf{Detection Cases analysis.} Finally, we present some of our detection results. Fig. \ref{fig7} shows two cases that our Group Detector successfully localize all the target groups with highly matched bounding boxes. In these two examples, it can be seen that our model achieve a high accuracy of predicting diversified UI components with different patterns and elements number as we discussed in Section \ref{sec1}.  

We also present some cases where our model fails. Fig. \ref{fig6} shows two typical examples that fully embody the existing shortcomings. The picture above shows that all group predictions are correct but not perfect fits. With an empty background around, our model fails to determine the boundaries of the target groups. Although this does not affect the localization of the layers in design drafts (no extra layers are included by error), the model performance is underestimated because of the low IoU. Another challenge is that our model shows weakness in distinguishing between foreground and background. As the bottom picture shows, our model mistakenly groups the address ``ARK'' in the background layer with the black dots in the foreground. The classic expression of this problem is that our model produces the wrong groups when faced with complex multi-level structures.  

\section{Conclusion}\label{sec5}
This paper investigates a novel issue about layers grouping in an automatic design draft to UI view code process, which can decrease the quality of generated code. To solve this issue, we propose our UI Layers Group Detector to locate the group accurately. By dataset segmentation, we achieve about a 20\% boost in detection AP. We also propose two plug-in components to help increase the detection performance. The Text fusion introduces text sematic and location prior knowledge and achieves about a 3.3\% increase in detection AP. For the box attention, the spatial binary images encoding potential text location give us a 2.5\% increase in detection AP.

\section*{Acknowledgement}
This research is supported by Alibaba-Zhejiang University Joint Research Institute of Frontier Technologies.



%
%
%
%
%
\bibliographystyle{splncs04}
\bibliography{references}






\end{document}